\documentclass[11pt]{article}

\usepackage[final]{acl}

\usepackage{times}
\usepackage{latexsym}

\usepackage[T1]{fontenc}

\usepackage[utf8]{inputenc}

\usepackage{microtype}

\usepackage{inconsolata}

\usepackage{graphicx}

\usepackage{multirow,array, balance, subcaption, adjustbox, booktabs}
\newcommand{\PreserveBackslash}[1]{\let\temp=\\#1\let\\=\temp}
\newcolumntype{C}[1]{>{\PreserveBackslash\centering}p{#1}}
\newcolumntype{R}[1]{>{\PreserveBackslash\raggedleft}p{#1}}
\newcolumntype{L}[1]{>{\raggedright\let\newline\\\arraybackslash\hspace{0pt}}m{#1}}

\title{Conditional Information Retrieval for Longitudinal Survey Harmonisation}

\author{Wing Yan Li \\ University of Surrey \\ \texttt{justina.li.106@gmail.com} \\\And
  Zeqiang Wang \\ University of Surrey \\ \texttt{zeqiang.wang@surrey.ac.uk} \\ \And
  Michelle Lin \\ University of Surrey \\ \texttt{linvmichelle@gmail.com} \\ \AND
  Jon Johnson \\ University College London \\ \texttt{jon.johnson@ucl.ac.uk} \\ \And
  Suparna De \\ University of Surrey \\ \texttt{s.de@surrey.ac.uk}
  }

\begin{document}
\maketitle
\begin{abstract}
Identifying conceptually equivalent questions across different surveys is a challenging information retrieval task where surface-level textual similarity is insufficient.
Two questions may be lexically similar yet assess entirely different sociological sub-concepts, or lexically different yet conceptually equivalent.
We introduce a conditional information retrieval framework that addresses this by decomposing the task into two steps: concept extraction, which identifies the underlying sociological concept of a survey question using a fine-tuned generative model, and concept-aware retrieval, which augments question representations with the extracted concept as an explicit conditioning signal for finer-grained alignment. 
To reflect the realistic data scarcity conditions of specialised social science annotation, we use CLOSER as a case study.
Our framework consistently outperforms baselines by 2–3 points in both MAP and Recall. 
We further analyse fine-tuning strategies including SFT, DPO, and dynamic DPO, finding that SFT delivers the most stable gains while dynamic DPO interacts differently with instruction-tuned versus distilled reasoning backbones. 
Finally, we demonstrate that a model fine-tuned on a general social science thesaurus (ELSST) transfers robustly to task-specific survey data, offering a practical solution under data scarcity.
\end{abstract}

\section{Introduction}

Sociological datasets such as longitudinal survey questionnaires are a rich source for studying the collective context and analysing societal shifts, serving as a rich evidence base for researchers and policymakers. Access to relevant and reliable data that can inform policy requires an understanding of the core \textit{sociological concept}. 
However, realising this potential requires data harmonisation, the identification of conceptually equivalent questions across different surveys. This is fundamentally a concept-conditioned retrieval problem: two questions are equivalent not because they share surface form, but because they assess the same underlying sociological construct. 
As the volume of archived survey data grows, manual annotation by domain experts becomes a critical bottleneck~\cite{li-etal-2025-are,o-neill-etal-2019}, motivating automated approaches that can reason about conceptual equivalence rather than surface-level textual similarity.

The core challenge is the inconsistent representation of theoretical constructs (e.g.,~concepts and sub-concepts), in which the same underlying sociological concept can be expressed in different wordings, response scales, or structural formats.
For example, \textit{What is he allergic to? Bee stings.} and \textit{Is he allergic to cats?}
While they might look similar on the surface, they do not have the same sociological concept.
The former has the concept of \textsc{Allergies; bee stings}, and the latter is \textsc{Allergies; cats}.
Conventional retrieval methods often fail in such cases because the critical difference lies in the sub-concept, which is signalled by subtle changes in the question text.~\citet{li-etal-2025-are} empirically confirmed that surface-level text similarity often fails to capture deeper conceptual equivalence, particularly when questions differ in fine-grained sub-concepts.
This reveals a critical research gap: conventional information retrieval models lack an explicit mechanism to capture the sociological concepts that underpin survey questions.

We address this gap by introducing a conditional information retrieval framework that explicitly grounds retrieval in sociological conceptual conditioning. 
Inspired by Conditional Semantic Textual Similarity~\cite{deshpande-etal-2023-c}, which demonstrated that similarity is fundamentally ambiguous without an explicit comparison dimension. 
Our key methodological insight is to decompose harmonisation into two steps: 
1. concept extraction: identifying the sociological concept of a given survey question using a fine-tuned generative model. 
2. concept-aware retrieval: augmenting question representations with the extracted concept as an explicit conditioning signal, enabling finer-grained alignment between questions that share conceptual meaning despite lexical or structural differences. 

We demonstrate our framework on CLOSER, an expert-annotated longitudinal survey dataset, which we treat as a case study reflecting the realistic data scarcity conditions of specialised social science annotation. Our empirical results show consistent improvements of 2–3 points in MAP and Recall over our baselines. Beyond the retrieval gains, we make the following contributions:

\begin{enumerate}
    \item A conditioned information retrieval pipeline for domain-specific data harmonisation, showing that decomposing retrieval into explicit concept extraction and concept-aware comparison improves both in-distribution and out-of-distribution performance on expert-annotated survey data.

    \item An empirical analysis of fine-tuning strategies: SFT, DPO, and dynamic DPO. Showing that SFT delivers the most consistent gains, and that dynamic DPO interacts differently with instruction-tuned versus distilled reasoning backbones, motivating further investigation into preference optimisation under domain-specific low-resource conditions.

    \item A cross-domain transfer study, demonstrating the effectiveness of applying a model fine-tuned on a general social science thesaurus (ELSST) \footnote{https://elsst.cessda.eu/} to our target survey datasets. This shows that robust sociological concept extraction is transferable across different conceptual vocabularies and data schemas, offering a practical solution when domain-specific labelled data is scarce.
\end{enumerate}

\section{Related Work}
\label{sec:rw}

\paragraph{Data Harmonisation.}
Longitudinal survey research generates rich datasets that span decades, countries, and demographic groups. Realising their comparative potential requires data harmonisation.
Data harmonisation is particularly important in longitudinal studies, as it enables meaningful comparisons across studies with respect to the underlying theoretical construct, despite differences in wording, response format, or structural context.
As the volume of archived survey data grows, manual annotation by domain experts becomes a critical bottleneck~\cite{li-etal-2025-are,o-neill-etal-2019}. 

Recent NLP work has begun to formalise this as a computational problem.
~\citet{li-etal-2025-are} empirically demonstrate that existing neural IR models struggle when questions exhibit high lexical overlap but differ in fine-grained, lower-level concepts, arguing that a balance between syntactic and semantic signals is essential.
Later,~\citet{wang-etal-2025-beyond} addressed the structural limitations of standard similarity measures for conceptual comparison, arguing that conceptual relations are not binary but structured like broader, narrower, related, or equivalent.
They proposed a Riemannian hierarchical ranking framework to better separate relation types in the embedding space. 

While complementary to our work, their approach requires carefully predefined concept hierarchies and labelled relation types. This limits applicability in settings where such structured annotations are unavailable, precisely the low-resource condition that characterises most longitudinal survey archives.~\citet{wang-etal-2026-trustworthy} took a different direction and framed harmonisation as closed-set classification with calibrated uncertainty.
This requires a stable, well-sampled label inventory, a condition that does not hold when concept vocabularies are open and evolving. 
Across these works, a consistent finding is that surface-level text matching is insufficient; explicit conceptual grounding is needed. 
Our framework addresses this directly by conditioning retrieval on automatically extracted sociological concepts, without requiring predefined hierarchies or stable label inventories.

\paragraph{Dense and Conditional Retrieval.}
Lately, neural information retrieval has shifted toward dense retrieval approaches that encode queries and documents into shared embedding spaces for efficient similarity search~\cite{karpukhin-etal-2020-dense}.
 Late interaction models such as ColBERT~\cite{khattab-etal-2020-colbert} and Poly-encoders~\cite{Humeau-etal-2020-Poly} extend this by preserving token-level interactions at retrieval time, offering a balance between expressiveness and efficiency. However, these approaches treat retrieval as unconditional and rank candidates without explicit grounding in the conceptual dimension along which equivalence should be assessed~\cite{li-etal-2025-are}.

While information is fundamentally a textual similarity task, this limitation motivates the idea of Conditional Semantic Textual Similarity (C-STS;~\cite{deshpande-etal-2023-c}), which demonstrated that similarity is fundamentally ambiguous without an explicit comparison dimension.~\citet{tu-etal-2024-linguistically} extended C-STS by anchoring conditions in typed linguistic features, improving interpretability and consistency. We build on this line of work by operationalising conditional similarity for structured, domain-specific vocabularies, and crucially, by extracting conditions automatically from the survey questions themselves rather than relying on manual annotation.

\paragraph{Model Fine-tuning.}
Supervised Fine-Tuning (SFT) is the standard approach for adapting pretrained language models to downstream tasks~\cite{ouyang-etal-2022-sft}.
Direct Preference Optimisation (DPO)~\cite{rafailov-etal-2023-dpo} extends this by training on pairs of preferred and rejected outputs without a separate reward model, though it is sensitive to the quality and balance of preference pairs~\cite{gee-etal-2025-code}.
Hence, they introduced dynamic sampling of preference pairs during training, demonstrating improved robustness and reduced overfitting in code generation. We adapt this strategy to the concept extraction as an ablation study.

\section{Conditional Information Retrieval}
\label{sec:framework}

\begin{figure*}[t!]
\centering
    \includegraphics[width=\textwidth]{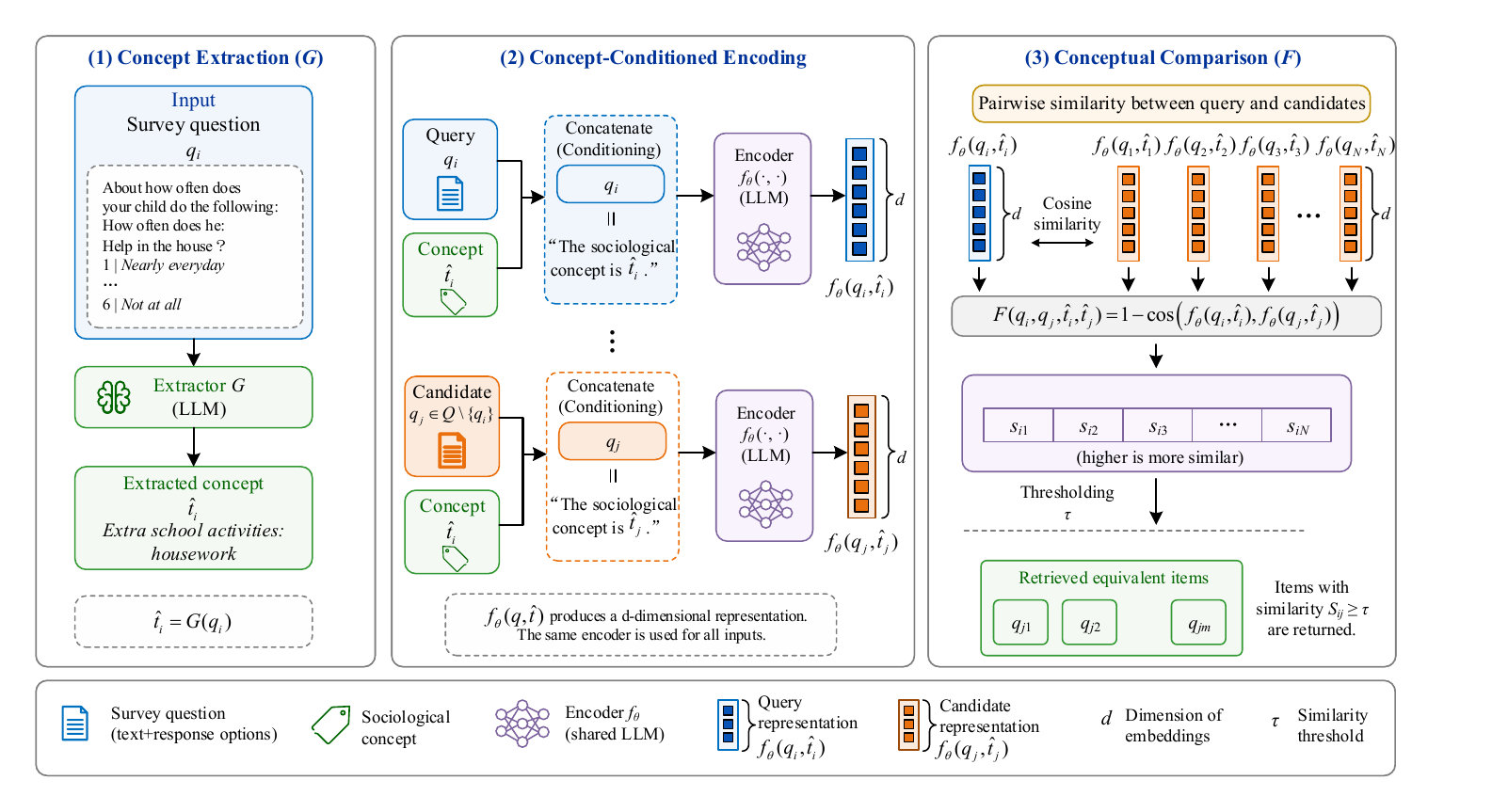}
    \caption{Two-step conditional retrieval framework for retrieving conceptually equivalent items. Given a survey question $q_i$, (1) Concept Extraction uses an LLM to extract the underlying sociological concept $\hat{t}_i$; (2) Concept-Conditioned Encoding concatenates the extracted concept with the question as a conditioning signal before encoding; (3) Conceptual Comparison computes pairwise cosine similarity between query and candidate representations, returning items above threshold $\tau$}
\label{fig:pipeline}
\end{figure*}

We decompose the harmonisation task the harmonisation problem into two functions, as shown in Figure~\ref{fig:pipeline}: \textbf{Concept Extraction} ($G$) and \textbf{Concept Comparison} ($F$). 
The former identifies the sociological concept of a given question ($q \in Q$); the latter compares questions using the conceptual terms extracted from the first step as the condition. 
The same backbone model is used in both functions to ensure the framework is working within the same semantic space.
Our motivation is to improve models' sensitivity towards questions' sociological concept and their nuanced relationships while maintaining efficient computation.

\paragraph{Concept Extraction ($G$)}
We use decoder-only models for sociological conceptual term extraction task, optimising $G(\cdot)$ using SFT and DPO. 
While SFT teaches models using preferred response only, DPO requires positive and negative contrastive samples for training. 
Given each question $q_i\in Q$ and the corresponding sociological conceptual term $t_i\in T$, negative samples are obtained by randomly selecting from the set of unique terms $T' = T \setminus \{t_i\}$ at the start of each training epoch.

\paragraph{Concept Comparison ($F$)}
To maintain the efficiency of the framework, we adopt the late interaction retrieval approach~\cite{khattab-etal-2020-colbert,Humeau-etal-2020-Poly,li-etal-2025-are} and leverage language models as a reader model for generating sequence representations. 
For each questions $q_i\in Q$ and $q_j\in Q \setminus \{q_i\}$, we extract the sociological conceptual terms with the optimised $G(\cdot)$ such that $\hat{t_i}=G(q_i)$ and $\hat{t_j}=G(q_j)$. 
We measure similarity as the cosine distance between the question pairs: 
\begin{equation}
    F( q_i, q_j, \hat{t_i}, \hat{t_j} ) = 1-cos(f_\theta(q_i,\hat{t_i}), f_\theta(q_j,\hat{t_j}))
\end{equation}
where $f_\theta{(\cdot)}$ computes the combined representation of the given question and conceptual term, and $cos$ is the cosine similarity. 
A threshold ($\tau$) is set at cosine distance $<$ 0.1.

\section{Experiments}

\begin{table}[t!]
\centering
\small
    \begin{tabular}{c|cc}
    \toprule
        Split & CLOSER-ID & CLOSER-OOD \\ \midrule
        Train & 611 & - \\ \midrule
        Test & 405 & 3479 \\
    \bottomrule
    \end{tabular}
\caption{Final count of instances per dataset. CLOSER-ID is deduplicated and split into train and test splits. CLOSER-OOD serves as the out-of-distribution test set only.}
\label{tab:data-stat}
\end{table}

\subsection{Datasets}

We use two longitudinal survey datasets drawn from the CLOSER project, a large-scale longitudinal social science survey project, tagged with sociological concepts by domain experts (survey data managers). 
\textbf{CLOSER-ID} contains 8,370 questions annotated with sociological concepts by domain experts.
We further clean the dataset by deduplicating on the (question $+$ response options, sociological conceptual term) pair before splitting, yielding 611 training and 405 test instances.
\textbf{CLOSER-OOD} contains 3,479 questions from a different set of survey sweeps that mimics real-world user data; instances whose inputs appear in CLOSER-Train are removed, giving a clean out-of-distribution test set.
The final count of each dataset and its splits is shown in Table~\ref{tab:data-stat}.

\subsection{Concept Extraction}

\paragraph{CLOSER Model.}
CLOSER models refer to models trained using the CLOSER-ID dataset. 
In this setting, we utilise DeepSeek-R1-Distill-Qwen-7B\footnote{https://huggingface.co/deepseek-ai/DeepSeek-R1-Distill-Qwen-7B} and Qwen2.5-7B-Instruct\footnote{https://huggingface.co/Qwen/Qwen2.5-7B-Instruct} as our base models. 
For training efficiency, we make use of Low-Rank Adaptation (LoRA;~\cite{hu-etal-2022-lora}) with rank = 32 and alpha = 64, applied to all the linear layers, a dropout of 0.1 is applied as well to reduce overfitting.
The prompt to align models for extracting the sociological conceptual term follows the format: ``\textit{Description of the input, [INPUT], task, guidelines and constraints.}'' (Appendix~\ref{appx:prompt}). 
Here, [INPUT] is the concatenation of the question and response options, if they exist, or that of the question and a single whitespace character otherwise.

\paragraph{ELSST Model.}
We also investigate the cross-domain transferability of a model fine-tuned on the synthetic European Language Social Science Thesaurus (ELSST) dataset~\cite{wang-etal-2025-beyond}, with Qwen2.5-7B-Instruct as the backbone model. 
This dataset consists of ELSST concepts and their definitions, which~\cite{wang-etal-2025-beyond} augmented by generating textual variations conditioned on different contexts and personas. 
We follow the same training setup as the SFT variants of the CLOSER models.
Given the discrepancy in the input-output schemas between the CLOSER and ELSST, the underlying knowledge representations learned by each model are inherently different. 
Hence, a performance drop when applying the ELSST model to CLOSER data is expected.

\subsection{Concept Comparison}

As mentioned, given $q$ and $\hat{t}$, both features are combined with $f_\theta{(\cdot)}$ for similarity calculation. 
Although the task is formulated as a retrieval problem, it is inherently a text similarity task. 
To reduce the potential ambiguity~\cite{deshpande-etal-2023-c} and address the granularity mismatch identified by~\cite{li-etal-2025-are}, we construct $f_\theta{(\cdot)}$ to produce conditioned representations. 
Conditions are written following the format: ``The sociological concept is [$\hat{t}$]'', a relatively well-defined condition~\cite{tu-etal-2024-linguistically}. 
Input to $f_\theta{(\cdot)}$ are the concatenations of $q$ and the corresponding condition.
We use the LoRA fine-tuned models from the concept extraction step as the reader model to obtain the embeddings.
We encode $q$ and $\hat{t}$ jointly rather than separately, as separate encoding may not provide sufficient contextual grounding for $\hat{t}$  to influence the resulting representation.
This a concern consistent with the encoding configuration findings in C-STS~\cite{deshpande-etal-2023-c}. 
Encoding them independently (e.g.,~via averaging) may not accurately reflect $\hat{t}$'s intended semantic role within the input.

\subsection{Evaluation}

\paragraph{Baselines.}
Our baselines are the conventional end-to-end retrieval model that do not consider any additional information.
Following the late interaction retrieval approach, we utilise the pre-trained DeepSeek-R1-Distill-Qwen-7B and Qwen2.5-7B-Instruct as our reader models. 
The output of each reader model is mean-pooled to generate sequence representations for ranking via cosine distance.

\paragraph{Metrics.}
We evaluate using Mean Average Precision at K (MAP@K) and Recall at K (Recall@K), where $K\in\{1, 5, 10\}$. 
A retrieved item is considered relevant if its ground truth conceptual term matches that of the query.
Together, these two metrics offer a comprehensive evaluation of our framework's effectiveness, balancing the aspects of ranking precision and retrieval coverage.

\section{Results}

\begin{table*}[t]
\centering
    \begin{subtable}[t]{\textwidth}
    \centering
    \small
        \begin{tabular}{c|c|ccc|ccc}
        \toprule
            Model & Variant & MAP@1 & MAP@5 & MAP@10 & Recall@1 & Recall@5 & Recall@10 \\ \midrule
             \multirow{5}{*}{\shortstack{Qwen2.5-7B-\\Instruct}} & Baseline & 67.99 & 71.69 & 71.68 & 37.40 & 64.00 & 70.59 \\ \cmidrule{2-8}
            & Pre-trained & 69.73 & 73.10 & 72.99 & 38.87 & 64.81 & \textbf{71.91} \\
            & SFT & \textbf{70.47} & \textbf{74.09} & \textbf{74.17} & \textbf{40.03} & \textbf{65.10} & 70.84 \\
            & DPO & 66.50 & 70.23 & 69.91 & 37.82 & 61.68 & 68.11 \\
            & ELSST & 69.48 & 72.78 & 72.65 & 38.81 & 65.45 & 71.54 \\ \midrule
            \multirow{4}{*}{\shortstack{DeepSeek-R1-\\Distill-Qwen-7B}} & Baseline & 70.47 & 73.97 & 73.92 & 39.06 & 66.15 & 74.25 \\ \cmidrule{2-8}
            & Pre-trained & 70.47 & 74.73 & 74.39 & 39.18 & 67.66 & 75.72 \\
            & SFT & \textbf{73.70} & \textbf{77.37} & \textbf{76.98} & \textbf{41.54} & \textbf{69.23} & \textbf{76.19} \\
            & DPO & 72.70 & 75.07 & 75.64 & 41.09 & 68.45 & 75.65 \\
        \bottomrule
        \end{tabular}
     \caption{CLOSER-ID (In-distribution).}
    \label{tab:v2-results}
    \end{subtable} \smallskip %
    
    \begin{subtable}[t]{\textwidth}
        \centering
        \small
        \begin{tabular}{c|c|ccc|ccc}
        \toprule
            Model & Variant & MAP@1 & MAP@5 & MAP@10 & Recall@1 & Recall@5 & Recall@10 \\ \midrule
            \multirow{5}{*}{\shortstack{Qwen2.5-7B-\\Instruct}} & Baseline & 70.32 & 73.33 & 72.06 & 17.78 & 32.44 & 40.43 \\ \cmidrule{2-8}
            & Pre-trained &  \textbf{70.81} & \textbf{73.93} & 72.71 & 17.83 & \textbf{33.09} & \textbf{41.26} \\
            & SFT & 70.72 & 73.57 & \textbf{72.77} & \textbf{17.92} & 32.78 & 40.67 \\
            & DPO & 70.00 & 72.57 & 71.58 & 17.51 & 31.93 & 39.18 \\
            & ELSST & 69.83 & 73.15 & 71.94 & 17.22 & 32.56 & 40.63 \\ \midrule
            \multirow{4}{*}{\shortstack{DeepSeek-R1-\\Distill-Qwen-7B}} & Baseline & 70.43 & 73.22 & 72.23 & 17.86 & 32.40 & 40.00 \\ \cmidrule{2-8}
            & Pre-trained & 69.59 & 72.92 & 71.74 & 17.31 & 32.76 & 40.87 \\
            & SFT & \textbf{70.96} & \textbf{73.95} & 72.62 & \textbf{17.98} & \textbf{33.07} & \textbf{41.39} \\
            & DPO & 70.70 & 73.78 & \textbf{72.66} & 17.78 & 32.94 & 40.97 \\
        \bottomrule
        \end{tabular}
    \caption{CLOSER-OOD (Out-of-Distribution).}
    \label{tab:v3-results}
    \end{subtable}

\caption{Results on both CLOSER-ID and CLOSER-OOD. Highest metric within each backbone model group is highlighted in \textbf{bold}. The ELSST model uses Qwen2.5-7B-Instruct as the backbone model. Note that other than the baseline, as it is not applicable, all models use the conditioned representations.}
\label{tab:results}
\end{table*}

Overall, our conditional information retrieval framework consistently outperformed the baseline retrieval models across the two datasets and backbone architectures.
On CLOSER-ID, the SFT-fine-tuned DeepSeek-R1-Distill-Qwen-7B model (R1-distill-SFT) has the highest scores.
MAP@10 reaches 76.98 and Recall@10 reaches 76.19, which are gains of 3.01 and 1.94 over their respective baselines.
Similarly, for Qwen2.5-7B-Instruct, the SFT variant (Instruct-SFT) achieves the best MAP@10 and Recall@10, outperforming the baseline by 2.95 and 1.32 points respectively.
Across the board, there is a substantial difference in Recall between K=1 and K=5.
For example, R1-distill-SFT shows a 27-point jump from 41.54 (Recall@1) to 69.23 (Recall@5).
Whilst there is a good balance between MAP@10 and Recall@10, this gap suggests that models do not always manage to retrieve relevant questions first but reliably retrieve conceptually equivalent items within the top-five results. 
This is practically meaningful for harmonisation workflows, where curators typically review a shortlist of candidates rather than a single result.
This suggest that our framework reliably surfaces equivalent questions within a reviewable set even when top-ranked precision is imperfect.
These results underscore the effectiveness of sociological conceptual conditioning in improving retrieval accuracy within a known distribution.

Evaluation on the out-of-distribution dataset (CLOSER-OOD) confirms the robustness and generalisation capability of our framework.
The performance gains, as expected, are subtler than those observed in CLOSER-ID, given the distributional shift.
Improvements remain consistent where conditioned retrieval models match or exceed the corresponding baselines across nearly all metrics.
A notable trend is the narrowing performance gap between SFT and DPO variants in this OOD setting.
While SFT fine-tuned models retain a slight edge, DPO variants exhibit less relative degradation compared to the performances on CLOSER-ID.
This suggests that the preference optimisation objective does offer some robustness to distributional shifts, yet not enough to surpass SFT.

The results further illustrate the cross-domain generalisation through the performance of the ELSST model.
Differing in both survey source and conceptual schema in the training data, both CLOSER-ID and CLOSER-OOD represent a distributional shift for the ELSST model. 
Despite this compounded distributional shift, the ELSST model achieves notably robust performance, attaining MAP@10 of 72.65 on CLOSER-ID, which is just 1.52 points below the best CLOSER model that has the same Qwen2.5-7B-Instruct backbone. 
Its competitive results on CLOSER-OOD confirm the feasibility as a practical, resource-efficient tool for conceptual retrieval when specialised labelled data is unavailable. 
This shows that the core task of sociological concept extraction is learnable and transferable across different conceptual vocabularies and data sources.

Finally, the comparison between fine-tuning objectives reveals distinct profiles, SFT provides stable, high-performance gains, while DPO exhibits greater variability. 
DPO sometimes narrows the performance gap with SFT in OOD settings (Table~\ref{tab:v3-results}), but underperforms more noticeably on CLOSER-OOD (Table~\ref{tab:v2-results}).
This suggests supervised fine-tuning is more effective for direct optimisation on a target distribution. 
Whereas the generalisation benefits of preference optimisation can be less predictable and involve trade-offs in ranking consistency.

\subsection{Are they really answering?}

\begin{table}[t!]
\centering
\begin{adjustbox}{width=\columnwidth}
    \begin{tabular}{c|c|cc}
    \toprule
        Model & Variant & CLOSER-ID & CLOSER-OOD \\ \midrule
        \multirow{4}{*}{\shortstack{Qwen2.5-7B-\\Instruct}} & Pre-trained & 52 & 492 \\
        & SFT & 0 & 0 \\
        & DPO & 1 & 17 \\
        & ELSST & 204 & 1759 \\ \midrule
        \multirow{3}{*}{\shortstack{DeepSeek-R1-\\Distill-Qwen-7B}} & Pre-trained & 0 & 11 \\
        & SFT & 0 & 0 \\
        & DPO & 1 & 9 \\
    \bottomrule
    \end{tabular}
\end{adjustbox}
\caption{Number of cases where the model is not returning any conceptual term. These could be cases where the thinking is unfinished or returning empty strings.}
\label{tab:nan-case}
\end{table}

A critical aspect of our framework is the reliability of the function G.
If it fails to produce a valid conceptual term, function F then lacks the necessary conditioning input, and the framework will involuntarily regress to the standard end-to-end ranking (i.e.,~the baseline).
Table~\ref{tab:nan-case} shows the frequency of empty extractions across different model variants.

SFT variants demonstrate better reliability than other variants for both backbone architectures.
This aligns with their superior quantitative performance (Table~\ref{tab:results}) and confirms that effective concept extraction is a foundation for the gains achieved via sociological conceptual conditioning.
In contrast, DPO variants exhibit moderate failure rates, consistent with their performance variability seen earlier.

The ELSST model shows the most pronounced failure rate, with 50.69\% of instances resulting in empty extractions on CLOSER-OOD.
Plausibly, this is due to the compounded schema and domain shift, highlighting a practical limitation.
This explains why its overall performance, while competitive, does not surpass CLOSER models. 
Pre-trained models show relatively low failure rates on CLOSER-ID but sharp increases on CLOSER-OOD, confirming that zero-shot concept extraction is fragile under domain shift and that fine-tuning is necessary for robustness.

\subsection{What are they returning?}
\label{subsec:quali2}
The core signal used for conditioning is the conceptual terms generated by G.
Table~\ref{tab:sample_concept-v2} in the Appendix illustrates a precision-sensitive test case requiring the model to isolate a single correct match from conceptually adjacent candidates. 
We report the top-three retrieved items to reveal the conceptual structure induced by different conditioning strategies and expose systematic semantic confusions.

For our example query in Table~\ref{tab:sample_concept-v2}, about household help frequency (ground-truth: \textit{{Extra school activities: housework}}), Instruct-SFT predicts the exact ground-truth format (\textit{Extra school activities: sports activities; swimming}), suggesting memorisation.
Yet, it fails to find the correct equivalent item, as the ranked results have mismatched sub-concepts.
This shows that memorisation, without understanding the sociological construct, remains susceptible to lexical overlap in the question text.

In contrast, R1-Distill-SFT and DPO variants produce paraphrases of the ground-truth term rather than exact reproductions.
For example, \textit{household chores} and \textit{family involvement}. Arguably, these models are better at capturing the sub-concept, a nuance that~\cite{li-etal-2025-are} found challenging.
This finer-grained alignment proves more effective for correct retrieval in precision-sensitive cases.

Pre-trained and ELSST models operate at a coarser conceptual level, outputting generic or thesaurus-driven terms that sacrifice fine-grained equivalence.
This pattern persists on CLOSER-OOD (Table~\ref{tab:sample_concept-v3}, Appendix).
Fine-tuned models consistently return more meaningful conceptual terms than pre-trained ones, further confirming that fine-tuning is necessary for nuanced conceptual comparison even under distributional shift.

\section{Ablation Study}

\begin{table*}[t]
\centering
    \begin{subtable}{\textwidth}
    \centering
    \small
        \begin{tabular}{c|ccc|ccc}
        \toprule
            Model & MAP@1 & MAP@5 & MAP@10 & Recall@1 & Recall@5 & Recall@10 \\ \midrule
            \multicolumn{7}{l}{CLOSER-ID (In-distribution)} \\ \midrule\midrule
            Qwen2.5-7B-Instruct & 69.98\% & 73.28\% & 73.49\% & 39.22\% & 64.70\% & 72.20\% \\
            DeepSeek-R1-Distill-Qwen-7B & 72.46\% & 75.61\% & 75.22\% & 41.09\% & 67.12\% & 75.94\% \\ \midrule
            \multicolumn{7}{l}{CLOSER-OOD (Out-of-Distribution)} \\ \midrule\midrule
            Qwen2.5-7B-Instruct & 70.29\% & 73.41\% & 72.39\% & 17.51\% & 32.86\% & 40.96\% \\
            DeepSeek-R1-Distill-Qwen-7B & 70.58\% & 73.54\% & 72.28\% & 17.56\% & 32.91\% & 41.09\% \\
        \bottomrule
        \end{tabular}
    \caption{Quantitative results of dynamic DPO models on CLOSER-ID and CLOSER-OOD.}
    \label{tab:dyn-dpo-quant}
    \end{subtable} \smallskip %
    
    \begin{adjustbox}{width=\textwidth}
    \begin{subtable}{\textwidth}
    \centering
    \small
         \begin{tabular}{C{0.13\textwidth}|C{0.15\textwidth}|L{0.22\textwidth}L{0.18\textwidth}L{0.18\textwidth}}
        \toprule
            \multirow{3}{*}{Model} & Predicted Concept for Query Question & \multirow{3}{*}{1} & \multirow{3}{*}{2} & \multirow{3}{*}{3} \\ \midrule
            
            \multirow{3}{*}{\shortstack{Qwen2.5-7B-\\Instruct}} & \multirow{3}{*}{\textbf{Helping}} & \textbf{Childhood activity frequency} & \textbf{Childhood} & \textbf{Helping} \\ \cmidrule{3-5}
             & & \textit{Extra school activities: sports activities; swimming} & \textit{Extra school activities: visit grandparents} & \textit{Extra school activities: housework} \\ \midrule
             
             \multirow{3}{*}{\shortstack{DeepSeek-R1-\\Distill-Qwen-7B}} & \multirow{3}{*}{\textbf{Frequency}} & \textbf{standardized questionnaire} & \textbf{Social Frequency} & \textbf{frequency of attendance} \\ \cmidrule{3-5}
             & & \textit{Extra school activities: sports activities; swimming} & \textit{Extra school activities: visit grandparents} & \textit{Extra school activities: religious} \\
        \bottomrule
        \end{tabular}
    \caption{Qualitative examples using the same query question as in Table~\ref{tab:sample_concept-v2} in Appendix with extracted conceptual terms are highlighted in \textbf{bold} and ground-truth conceptual terms in \textit{italics}.}
    \label{tab:dyn-dpo-sample}
    \end{subtable}
    \end{adjustbox}
\caption{Ablation study results obtained using models trained with dynamic DPO training. Quantitative results (a) and qualitative examples (b) illustrate the impact of dynamic sampling on metrics and conceptual term granularity.}
\label{tab:dyn-dpo}
\end{table*}

As mentioned in Section~\ref{sec:framework}, we randomly sample the negative samples at the \textit{beginning} of each training loop when training the DPO variants.
While the conventional sampling method shows that the DPO variants are performing less stably than the SFT variants, we utilise the dynamic sampling as suggested in Gee et al. (2025) (Section~\ref{sec:rw}), referred to as dynamic DPO. 
Under this setting, models will select the negative sample per $q_i$ \textit{after entering} the training loop.
Not only does this augment the data, but it is also expected that the model will increase the robustness from using static negatives.
Table~\ref{tab:dyn-dpo} shows both quantitative and qualitative results of models trained with dynamic DPO.

Looking at Table~\ref{tab:dyn-dpo-quant}, dynamic DPO outperforms conventional DPO with the Qwen2.5‑7B‑Instruct model on both datasets by having the biggest improvement on CLOSER-ID with 3.58 points increase in MAP@10, and a 4.09 points increase in Recall@10.
However, DeepSeek‑R1‑Distill‑Qwen‑7B model shows no improvement and exhibits a mild degradation, with an average drop of 0.3 points on both datasets.
Overall, comparing Table 2 and Table 5, dynamic DPO does not outperform the SFT variants.

The inconsistent impact of dynamic sampling across different backbone models is likely attributable to the DeepSeek‑R1‑Distilled models' differing thinking processes at inference time.
Unlike instruction tuned models that are optimised to follow varied prompts and adapt dynamically, DeepSeek‑R1‑Distilled models are specifically trained on chain-of-thought reasoning. 
Crucially, this reasoning property is ``locked'' over the training process.
Therefore, when dynamic negative sampling is applied, this introduces variability that might disrupt the DeepSeek‑R1‑Distilled models' previously learned reasoning chains, whereas the Qwen2.5‑7B‑Instruct model treats this variability as helpful data augmentation.

We further inspect the conceptual terms predicted by dynamic DPO using the challenging query from Section~\ref{subsec:quali2} (Table~\ref{tab:sample_concept-v2}). 
As shown in Table~\ref{tab:dyn-dpo-sample}, dynamic DPO yields more abstract conceptual terms for both models compared to conventional DPO.
For DeepSeek‑R1‑Distill‑Qwen‑7B, this shows a shift away from precise sub-concept alignments. 
The corresponding drop in granularity aligns with the performance degradation in Table~\ref{tab:dyn-dpo-quant}, supporting the hypothesis that dynamic DPO disrupts its structured reasoning pathways.
However, for Qwen2.5‑7B‑Instruct, this abstraction corresponds to its performance gains, suggesting the variability acts as beneficial data augmentation that encourages more generalisable representations.
Altogether, dynamic DPO trades fine‑grained conceptual precision for broader generalisation, a trade-off that benefits the adaptable instruct model but undermines the distilled reasoning model.

\section{Conclusion}

In this work, we introduced a conditional information retrieval framework for longitudinal survey harmonisation that decomposes the task into concept extraction and concept-aware retrieval.
Experiments on both in-distribution (CLOSER-ID) and out-of-distribution (CLOSER-OOD) datasets demonstrate consistent improvements of 2-3 points in MAP and Recall over baseline models. 
We found that supervised fine-tuning (SFT) variants deliver consistent performance gains, while DPO variants offer some robustness to distributional shifts but do not surpass SFT.
Dynamic DPO does not yield consistent improvements, with its effectiveness varying by backbone architecture. 
It is beneficial to the instruction-tuned models while degrading performance in distilled reasoning models. 
The ELSST transfer results confirm that sociological concept extraction can generalise across conceptual vocabularies, offering a practical solution when domain-specific labelled data is scarce.

There are several directions to further extend this framework.
Firstly, the framework could be applied to other specialised domains where conceptual equivalence matters but expert-annotated data is scarce, such as medical or clinical survey harmonisation, domains that use complex and specialised language. 
Secondly, the automatic concept extraction step currently relies on a single generated term; exploring multi-granular or set-valued concept representations could better capture questions that span multiple sociological constructs simultaneously. 
Thirdly, while we demonstrate transferability from ELSST to CLOSER, a more systematic investigation of cross-domain transfer, such as varying the source thesaurus, domain, and annotation style, would help establish the conditions under which transfer is reliable. 
Finally, the framework currently treats concept extraction and concept-aware retrieval as sequential steps; a jointly trained end-to-end model that optimises both steps simultaneously may yield further gains, particularly in low-resource settings where extraction errors compound retrieval errors.

\section*{Limitations}

This work has several limitations. First, our evaluation is conducted on CLOSER, a single expert-annotated longitudinal survey dataset, which we treat as a case study. 
While this reflects realistic data scarcity conditions in specialised social science annotation, the generalisability of our findings to other survey archives or annotation styles remains to be established. 
Second, results are reported from single runs; preliminary experiments suggested low variance across seeds, though formal significance testing remains future work.
Third, the concept extraction step relies on a single generated term per question; questions that span multiple sociological constructs simultaneously may not be adequately represented by this single-term conditioning signal. 
Fourth, our framework treats concept extraction and concept-aware retrieval as sequential steps, meaning extraction errors directly compound retrieval errors with no mechanism for recovery. 
Fifth, while we investigate two backbone architectures, the interaction between dynamic DPO and model training regime is observed on limited evidence, a more systematic investigation across architectures would be needed to draw stronger conclusions. 
Finally, the cosine distance threshold $\tau$ is set at 0.1 without ablation; its sensitivity to domain or dataset characteristics is unexplored.

\section*{Ethical Considerations}

This work uses existing, publicly available survey metadata.
Question text and expert-assigned sociological concepts does not involve the collection of new data or human subjects. 
No individual-level survey responses are used at any point. 
The CLOSER dataset is accessed under its standard research use conditions. 
The primary intended use of this framework is to assist domain experts in data harmonisation workflows, not to replace expert judgment. 
We do not foresee direct harms arising from this work, though we note that automated concept assignment in social science contexts should remain subject to expert review, as misclassification of sociological concepts could affect downstream comparative analyses.

\bibliography{custom}

@inproceedings{li-etal-2025-are,
    author = {Li, Wing Yan and Wang, Zeqiang and Johnson, Jon and De, Suparna},
    title = {Are Information Retrieval Approaches Good at Harmonising Longitudinal Surveys in Social Science?},
    year = {2025},
    isbn = {9798400715921},
    publisher = {Association for Computing Machinery},
    url = {https://doi.org/10.1145/3726302.3730164},
    doi = {10.1145/3726302.3730164},
    booktitle = {Proceedings of the 48th International ACM SIGIR Conference on Research and Development in Information Retrieval},
    location = {Padua, Italy},
    series = {SIGIR '25}
}

@inproceedings{wang-etal-2025-beyond,
    author = {Wang, Zeqiang and Li, Wing Yan and Johnson, Jon and Sastry, Nishanth and De, Suparna},
    title = {Beyond Surface Similarity: A Riemannian Hierarchical Ranking Framework for Sociological Concept Equivalence},
    year = {2025},
    isbn = {9798400720406},
    publisher = {Association for Computing Machinery},
    url = {https://doi.org/10.1145/3746252.3761029},
    doi = {10.1145/3746252.3761029},
    booktitle = {Proceedings of the 34th ACM International Conference on Information and Knowledge Management},
    location = {Seoul, Republic of Korea},
    series = {CIKM '25}
}

@inproceedings{hu-etal-2022-lora,
    title={Lo{RA}: Low-Rank Adaptation of Large Language Models},
    author={Edward J Hu and Yelong Shen and Phillip Wallis and Zeyuan Allen-Zhu and Yuanzhi Li and Shean Wang and Lu Wang and Weizhu Chen},
    booktitle={International Conference on Learning Representations},
    year={2022},
    url={https://openreview.net/forum?id=nZeVKeeFYf9}
}

@inproceedings{tu-etal-2024-linguistically,
    title = "Linguistically Conditioned Semantic Textual Similarity",
    author = "Tu, Jingxuan and Xu, Keer  and Yue, Liulu  and Ye, Bingyang  and Rim, Kyeongmin  and Pustejovsky, James",
    booktitle = "Proceedings of the 62nd Annual Meeting of the Association for Computational Linguistics (Volume 1: Long Papers)",
    year = "2024",
    address = "Bangkok, Thailand",
    publisher = "Association for Computational Linguistics",
    url = "https://aclanthology.org/2024.acl-long.64/",
    doi = "10.18653/v1/2024.acl-long.64",
}

@inproceedings{deshpande-etal-2023-c,
    title = "{C}-{STS}: Conditional Semantic Textual Similarity",
    author = "Deshpande, Ameet  and Jimenez, Carlos  and Chen, Howard  and Murahari, Vishvak  and Graf, Victoria  and Rajpurohit, Tanmay  and Kalyan, Ashwin  and Chen, Danqi  and Narasimhan, Karthik",
    booktitle = "Proceedings of the 2023 Conference on Empirical Methods in Natural Language Processing",
    year = "2023",
    address = "Singapore",
    publisher = "Association for Computational Linguistics",
    url = "https://aclanthology.org/2023.emnlp-main.345/",
    doi = "10.18653/v1/2023.emnlp-main.345",
}

@inproceedings{khattab-etal-2020-colbert,
    author = {Khattab, Omar and Zaharia, Matei},
    title = {ColBERT: Efficient and Effective Passage Search via Contextualized Late Interaction over BERT},
    year = {2020},
    booktitle = {Proceedings of the 43rd International ACM SIGIR Conference on Research and Development in Information Retrieval},
}

@inproceedings{Humeau-etal-2020-Poly,
    title={Poly-encoders: Architectures and Pre-training Strategies for Fast and Accurate Multi-sentence Scoring},
    author={Samuel Humeau and Kurt Shuster and Marie-Anne Lachaux and Jason Weston},
    booktitle={International Conference on Learning Representations},
    year={2020},
    url={https://openreview.net/forum?id=SkxgnnNFvH}
}

@inproceedings{rafailov-etal-2023-dpo,
author = {Rafailov, Rafael and Sharma, Archit and Mitchell, Eric and Ermon, Stefano and Manning, Christopher D. and Finn, Chelsea},
title = {Direct preference optimization: your language model is secretly a reward model},
year = {2023},
booktitle = {Proceedings of the 37th International Conference on Neural Information Processing Systems},
articleno = {2338},
series = {NIPS '23}
}

@inproceedings{gee-etal-2025-code,
    title = "Code-Optimise: Self-Generated Preference Data for Correctness and Efficiency",
    author = "Gee, Leonidas  and Gritta, Milan  and Lampouras, Gerasimos  and Iacobacci, Ignacio",
    booktitle = "Findings of the Association for Computational Linguistics: NAACL 2025",
    year = "2025",
    publisher = "Association for Computational Linguistics",
    url = "https://aclanthology.org/2025.findings-naacl.5/",
}

@inproceedings{ouyang-etal-2022-sft,
    author = {Ouyang, Long and Wu, Jeff and Jiang, Xu and Almeida, Diogo and Wainwright, Carroll L. and Mishkin, Pamela and Zhang, Chong and Agarwal, Sandhini and Slama, Katarina and Ray, Alex and Schulman, John and Hilton, Jacob and Kelton, Fraser and Miller, Luke and Simens, Maddie and Askell, Amanda and Welinder, Peter and Christiano, Paul and Leike, Jan and Lowe, Ryan},
    title = {Training language models to follow instructions with human feedback},
    year = {2022},
    isbn = {9781713871088},
    publisher = {Curran Associates Inc.},
    doi = {10.48550/arXiv.2203.02155},
    url = {https://doi.org/10.48550/arXiv.2203.02155},
    booktitle = {Proceedings of the 36th International Conference on Neural Information Processing Systems},
    articleno = {2011},
    location = {New Orleans, LA, USA},
    series = {NIPS '22}
}

@article{o-neill-etal-2019,
author = {O’Neill, Dara and Benzeval, Michaela and Boyd, Andy and Calderwood, Lisa and Cooper, Cyrus and Corti, Louise and Dennison, Elaine and Fitzsimons, Emla and Goodman, Alissa and Hardy, Rebecca and Inskip, Hazel and Molloy, Lynn and Sacker, Amanda and Sudlow, Allan and Sullivan, Alice and Park, Alison},
title = {Data Resource Profile: Cohort and Longitudinal Studies Enhancement Resources (CLOSER)},
journal = {International Journal of Epidemiology},
volume = {48},
number = {3},
pages = {675-676i},
year = {2019},
month = {06},
issn = {0300-5771},
doi = {10.1093/ije/dyz004},
url = {https://doi.org/10.1093/ije/dyz004},
eprint = {https://academic.oup.com/ije/article-pdf/48/3/675/29005340/dyz004.pdf},
}

@article{wang-etal-2026-trustworthy, 
title={Trustworthy Classification for Complex Social Surveys: A Memory-Enhanced Hierarchical Framework with Calibrated Uncertainty}, 
url={https://ojs.aaai.org/index.php/AAAI/article/view/41291}, 
DOI={10.1609/aaai.v40i46.41291}, 
number={46}, 
journal={Proceedings of the AAAI Conference on Artificial Intelligence}, 
author={Wang, Zeqiang and Oldroyd, Rebecca and Wang, Yuqi and Wu, Jiageng and Yang, Jie and Wang, Wei and Sastry, Nishanth R. and Johnson, Jon and De, Suparna}, 
year={2026}, 
month={Mar.},
pages={39415–39422} 
}

@inproceedings{karpukhin-etal-2020-dense,
    title = "Dense Passage Retrieval for Open-Domain Question Answering",
    author = "Karpukhin, Vladimir  and
      Oguz, Barlas  and
      Min, Sewon  and
      Lewis, Patrick  and
      Wu, Ledell  and
      Edunov, Sergey  and
      Chen, Danqi  and
      Yih, Wen-tau",
    editor = "Webber, Bonnie  and
      Cohn, Trevor  and
      He, Yulan  and
      Liu, Yang",
    booktitle = "Proceedings of the 2020 Conference on Empirical Methods in Natural Language Processing (EMNLP)",
    month = nov,
    year = "2020",
    address = "Online",
    publisher = "Association for Computational Linguistics",
    url = "https://aclanthology.org/2020.emnlp-main.550/",
    doi = "10.18653/v1/2020.emnlp-main.550",
    pages = "6769--6781",
}

\appendix
\section{Prompts}
\label{appx:prompt}
The prompt we used in our experiments is: \textit{``The input is a concatenation of a question and its corresponding response options. Identify the sociological conceptual term of the input: ``{x['QuestionText']} + {x['response\_domain']}''. ONLY return ONE conceptual term exactly in this format: |||{{concept}}|||.''} a \texttt{<think>} token is added at the end to trigger the thinking function when DeepSeek‑R1‑Distill‑Qwen‑7B is the backbone model, at inference time.

\begin{table*}[t]
\centering
\begin{adjustbox}{width=\textwidth}
    \begin{subtable}{\textwidth}
    \centering
    \small
         \begin{tabular}{C{0.1\textwidth}|C{0.15\textwidth}|L{0.2\textwidth}L{0.2\textwidth}L{0.2\textwidth}}
        \toprule
            \multirow{3}{*}{Variant} & Predicted Concept for Query Question & \multirow{3}{*}{1} & \multirow{3}{*}{2} & \multirow{3}{*}{3} \\ \midrule
            
            \multirow{4}{*}{Pre-trained} & \multirow{4}{*}{\textbf{Routine}} & \textbf{nan} & \textbf{frequency} & \textbf{nan} \\ \cmidrule{3-5}
            & & \textit{Extra school activities: sports activities; swimming} & \textit{Extra school activities: computer games} & \textit{Extra school activities: visit grandparents} \\ \midrule
            
            \multirow{4}{*}{SFT} & \multirow{4}{*}{\textbf{\shortstack{Extra school\\activities:\\housework}}} & \textbf{Extra school activities: sports activities; swimming} & \textbf{Extra school activities: computer games} & \textbf{Extra school activities: visit grandparents} \\ \cmidrule{3-5}
            & & \textit{Extra school activities: sports activities; swimming} & \textit{Extra school activities: computer games} & \textit{Extra school activities: visit grandparents} \\ \midrule
            
            \multirow{4}{*}{DPO} & \multirow{4}{*}{\textbf{Housework}} & \textbf{Swimming frequency} & \textbf{Religious education frequency} & \textbf{Frequency of family interaction} \\ \cmidrule{3-5}
            & & \textit{Extra school activities: sports activities; swimming} & \textit{Extra school activities: religious} & \textit{Extra school activities: visit grandparents} \\ \midrule
            
            \multirow{4}{*}{ELSST} & \multirow{4}{*}{\textbf{FREQUENCY}} & \textbf{FREQUENCY} & \textbf{FREQUENCY} & \textbf{nan} \\ \cmidrule{3-5}
            & & \textit{Extra school activities: sports activities; swimming} & \textit{Extra school activities: computer games} & \textit{Extra school activities: religious} \\
        \bottomrule
        \end{tabular}
    \caption{Qwen2.5-7B-Instruct}
    \end{subtable}
\end{adjustbox} \quad

\begin{adjustbox}{width=\textwidth}
      \begin{subtable}{\textwidth}
    \centering
    \small
         \begin{tabular}{C{0.1\textwidth}|C{0.15\textwidth}|L{0.2\textwidth}L{0.2\textwidth}L{0.2\textwidth}}
        \toprule
            \multirow{3}{*}{Variant} & Predicted Concept for Query Question & \multirow{3}{*}{1} & \multirow{3}{*}{2} & \multirow{3}{*}{3} \\ \midrule
            
            \multirow{4}{*}{Pre-trained} & \multirow{4}{*}{\textbf{Social Behavior}} & \textbf{Social Measurement} & \textbf{family involvement} & \textbf{Survey} \\ \cmidrule{3-5}
            & & \textit{Extra school activities: sports activities; swimming} & \textit{Extra school activities: visit grandparents} & \textit{Extra school activities: housework} \\ \midrule
            
            \multirow{4}{*}{SFT} & \multirow{4}{*}{\textbf{\shortstack{household\\responsibilities\\towards children}}} & \textbf{household chores} & \textbf{frequency} & \textbf{frequency of behavior} \\ \cmidrule{3-5}
            & & \textit{Extra school activities: housework} & \textit{Extra school activities: visit grandparents} & \textit{Extra school activities: sports activities; swimming} \\ \midrule
            
            \multirow{4}{*}{DPO} & \multirow{4}{*}{\textbf{\shortstack{Social Behavior\\Patterns}}} & \textbf{family involvement} & \textbf{Social participation} & \textbf{frequency} \\ \cmidrule{3-5}
            & & \textit{Extra school activities: visit grandparents} & \textit{Extra school activities: sports activities; swimming} & \textit{Extra school activities: housework} \\
        \bottomrule
        \end{tabular}
    \caption{DeepSeek-R1-Distill-Qwen-7B}
    \end{subtable}
\end{adjustbox}
\caption{Top-three retrieved items for the query ``\texttt{About how often does your child do the following: How often does he: Help in the house 1|Nearly every day. ... 6|Not at all}'' (ground-truth concept: \textit{Extra school activities: housework}). \textbf{Bold} entries show the model-predicted concept used for conditioning; \textit{italicised} entries show the annotated gold concept of each retrieved item.}
\label{tab:sample_concept-v2}
\end{table*}

\begin{table*}
\centering
    \begin{adjustbox}{width=\textwidth}
    \begin{subtable}{\textwidth}
    \centering
    \small
            \begin{tabular}{C{0.1\textwidth}|C{0.15\textwidth}|L{0.2\textwidth}L{0.2\textwidth}L{0.2\textwidth}}
            \toprule                
                \multirow{2}{*}{Variant} & Predicted Concept for Query Question & \multirow{2}{*}{1} & \multirow{2}{*}{2} & \multirow{2}{*}{3} \\ \midrule
                
                Pre-trained & \multirow{2}{*}{\textbf{lack of remorse}} & \textbf{disobedience}  & \textbf{Paranoia} & \textbf{temperament} \\ \cmidrule{3-5}
                & & \textit{childdev} & \textit{childdev} & \textit{childdev} \\ \midrule
                
                \multirow{2}{*}{SFT} & \multirow{2}{*}{\textbf{\shortstack{Malaise:\\post conflict}}} & \textbf{Malaise: temper} & \textbf{Malaise: persecuted} & \textbf{Malaise: attention seeking} \\ \cmidrule{3-5}
                & & \textit{childdev} &  \textit{childdev} &  \textit{childdev} \\ \midrule
                
                \multirow{2}{*}{DPO} & \multirow{2}{*}{\textbf{\shortstack{would be the\\answer.anmar:}}} & \textbf{persecution} & \textbf{feel||/, and love as} & \textbf{Disobedience} \\ \cmidrule{3-5}
                & & \textit{childdev} & \textit{childdev} & \textit{childdev} \\ \midrule
    
                \multirow{3}{*}{\shortstack{dynamic\\DPO}} & \multirow{3}{*}{\textbf{Emotion}} & \textbf{Disobedience} & \textbf{Temper} & \textbf{feeling of persecution} \\ \cmidrule{3-5}
                & & \textit{childdev} & \textit{childdev} & \textit{childdev} \\ \midrule
                
                \multirow{3}{*}{ELSST} & \multirow{3}{*}{\textbf{nan}} & \textbf{nan} & \textbf{nan} & \textbf{PARANOIA} \\ \cmidrule{3-5}
                & & \textit{childdev} & \textit{childdev} & \textit{childdev} \\
            \bottomrule
            \end{tabular}
    \caption{Qwen2.5-7B-Instruct.}
    \end{subtable} 
    \end{adjustbox}
    
    \begin{adjustbox}{width=\textwidth}
    \begin{subtable}{\textwidth}
    \centering
    \small
            \begin{tabular}{C{0.1\textwidth}|C{0.15\textwidth}|L{0.2\textwidth}L{0.21\textwidth}L{0.19\textwidth}}
            \toprule
                \multirow{2}{*}{Variant} & Predicted Concept for Query Question & \multirow{2}{*}{1} & \multirow{2}{*}{2} & \multirow{2}{*}{3} \\ \midrule
                
                \multirow{2}{*}{Pre-trained} & \multirow{3}{*}{\textbf{\shortstack{Social Bonding\\Theory}}} & \textbf{Social Phobia Scale} & \textbf{Normative Deviance} & \textbf{Social Control} \\ \cmidrule{3-5}
                & & \textit{childdev} & \textit{childdev} & \textit{childdev} \\ \midrule
                
                \multirow{2}{*}{SFT} & \multirow{2}{*}{\textbf{norms}} & \textbf{social bonding} & \textbf{social exclusion} & \textbf{attention-seeking behavior} \\ \cmidrule{3-5}
                & & \textit{childdev} & \textit{childdev} & \textit{childdev} \\ \midrule
                
                \multirow{2}{*}{DPO} & \multirow{2}{*}{\textbf{\shortstack{Moral\\Disengagement}}} & \textbf{Social comparison} & \textbf{Frequency of occurrence} & \textbf{Social validation} \\ \cmidrule{3-5}
                & & \textit{childdev} & \textit{childdev} & \textit{childdev} \\ \midrule
    
                \multirow{3}{*}{\shortstack{dynamic\\DPO}} & \multirow{3}{*}{\textbf{culpabilism}} & \textbf{Defensive Psychology} & \textbf{social status} & \textbf{Social bonding} \\ \cmidrule{3-5}
                & & \textit{childdev} & \textit{childdev} & \textit{childdev} \\
            \bottomrule
            \end{tabular}
    \caption{DeepSeek-R1-Distill-Qwen-7B.}
    \end{subtable}
    \end{adjustbox}
\caption{Top-three retrieved items for the query ''\texttt{He/She does not seem to feel sorry after he/she misbehaves. 1|Often true. 2|Sometimes true. 3|Not true}`` (ground-truth concept: \textit{childdev}). Extracted conceptual terms  are in \textbf{bold} and ground-truth conceptual terms are in \textit{italics}.}
\label{tab:sample_concept-v3}
\end{table*}

\end{document}